\documentclass[9pt,conference]{IEEEtran}
\ifCLASSINFOpdf
\else
\fi
\usepackage{amssymb}
\usepackage{amstext}
\usepackage{amsmath}
\usepackage{graphicx}

\usepackage{amsmath}

\DeclareMathOperator*{\argmin}{arg\,min}

\begin{document}
%
\title{Learning a Lie Algebra from Unlabeled Data Pairs}

\author{\IEEEauthorblockN{Christopher Ick and Vincent Lostanlen \\ Music and Audio Research Lab, New York University, New York, NY, USA}}


%


\maketitle

%
\IEEEpeerreviewmaketitle

\section{Introduction}


Disentangling factors of variability from natural data is a core principle of modern representation learning \cite{Deeplearning:Goodfellow}.
For example, in computer vision, small changes in illumination and viewpoint for a given object may incur large pixel-wise perturbations.
Therefore, robust models for object recognition must operate in a feature space of reduced dimensionality in which these perturbations are systematically canceled.
Deep convolutional networks (convnets) show a remarkable ability to learn disentangled representations \cite{Sliders:Lample}.
Despite this accomplishment, the mathematical foundations of representation learning in convnets is still in infancy \cite{DCNNs:Mallat}.

In recent years, the generalization of deep learning to Lie groups beyond rigid motion in $\mathbb{R}^n$ has allowed to build convnets over datasets with non-trivial symmetries \cite{Equivariance:Kondor}, such as patterns over the surface of a sphere \cite{SphericalCNNs:Cohen}.
However, one limitation of this approach is the need to explicitly define the Lie group underlying the desired invariance property before training the convnet.
Whereas rotations on the sphere have a well-known symmetry group ($\mathrm{SO}(3)$), the same cannot be said of many real-world factors of variability.
For example, in machine listening, pitch shifts induce a complex effect in the time--frequency domain by transposing the harmonic source while leaving the spectral envelope unchanged \cite{Invariance:Lostanlen}.
More generally, the disentanglement of pitch, intensity dynamics, and playing technique remains a challenging task in music information retrieval \cite{Technique:Lostanlen}.

This article proposes a machine learning method to discover a nonlinear transformation of the space $\mathbb{R}^n$ which maps a collection of $n$-dimensional vectors $(\boldsymbol{x}_i)_i$ onto a collection of target vectors $(\boldsymbol{y}_i)_i$.
The key idea is to approximate every target $\boldsymbol{y}_i$ by a matrix--vector product of the form $\boldsymbol{\widetilde{y}}_i = \boldsymbol{\phi}(t_i) \boldsymbol{x}_i$, where the matrix $\boldsymbol{\phi}(t_i)$ belongs to a one-parameter subgroup of $\mathrm{GL}_n (\mathbb{R})$, denoted by $G$ in the following.
Crucially, the value of the parameter $t_i \in \mathbb{R}$ may change between data pairs  $(\boldsymbol{x}_i, \boldsymbol{y}_i)$ and does not need to be known in advance.


\section{Theory}

By Stone's theorem \cite[chapter 10]{Lie:Hall}, there exists a matrix $\mathbf{A}$ such that $G$ can be described by matrix exponentiations of the form:
\begin{equation}
    G =
    \left\{
    \boldsymbol{\phi}(t) =
    \mathrm{e}^{t \mathbf{A}}
    =
    \sum_{k=0}^{+\infty}
    \dfrac{t^k}{k!} \mathbf{A}^k
    \;\Bigg\vert\; t\in\mathbb{R}
    \right\}.
\end{equation}
The matrix $\mathbf{A}$ is called an infinitesimal generator of the Lie group $G$, and its linear span $\mathfrak{g} = \{t\mathbf{A}\,\vert\, t\in\mathbb{R}\}$ is called the Lie algebra associated to $G$.
Furthermore, the application $\boldsymbol{\phi}$ defines a group homomorphism from $\mathbb{R}$ to $G$.
Evaluating the first derivative of this application near the identity ($t=0$) yields the matrix $\mathbf{A}$. Hence:
\begin{equation}
    \forall\boldsymbol{x}\in\mathbb{R}^n,\;
    \mathbf{A}\boldsymbol{x} =
    \lim_{t\rightarrow 0}
    \dfrac{
    \boldsymbol{\phi}(t)\boldsymbol{x}
    - \boldsymbol{x}
    }{t}
    =
    \left(
    \left.
    \frac{\mathrm{d}\boldsymbol{\phi}
    }{
    \mathrm{d}t}\right|_{t=0}\right)\boldsymbol{x}.
\end{equation}

The optimal parameter $t_i$ corresponding to each pair $(\boldsymbol{x}_i$, $\boldsymbol{y}_i$) can be assumed to be small ($\vert t_i \vert \ll 1$) but not infinitesimal.
We circumvent this issue by linearizing the orbit of $G$ near $\boldsymbol{x}_i$:
\begin{equation}
    \boldsymbol{y}_i \approx
    \widetilde{\boldsymbol{y}}_i =
    \boldsymbol{\phi}(t_i) \boldsymbol{x}_i =
    \mathrm{e}^{t_i \mathbf{A}} \boldsymbol{x}_i =
    \boldsymbol{x}_i +
    t_i \mathbf{A} \boldsymbol{x} +
    O(t_i^2).
    \label{eq:linapprox}
\end{equation}
It follows from the equation above that, as long as $t_i$ remains small, the vectors $(\boldsymbol{y}_i - \boldsymbol{x}_i)$ and ($\mathbf{A}\boldsymbol{x}_i$) are approximately colinear.
Therefore, we propose to discover $\mathbf{A}$ by minimizing the empirical risk associated to the rectified cosine distance between these vectors:
\begin{equation}
    \mathbf{A}^{\ast} =
    \argmin_{
    \mathbf{A}\in\mathbb{R}^{n\times n}
    }
    \sum_{i}
    \left(
    1 -
    \dfrac{
    \big\vert
    (\boldsymbol{y}_i
    - \boldsymbol{x}_i)^{\top}
    \mathbf{A}
    \boldsymbol{x}_i
    \big\vert
    }{
    \Vert\boldsymbol{y}_i
    - \boldsymbol{x}_i \Vert_2
    \Vert \mathbf{A}\boldsymbol{x}_i \Vert_2
    }
    \right).
    \label{eq:cosine-distance}
\end{equation}


\section{Practice}
We define $\mathbf{A}$ as the weight matrix of a feedforward neural network with a single layer and a linear activation function.
By gradient descent on the cosine distance loss (see Equation \ref{eq:cosine-distance}), we train $\mathbf{A}$ to predict the azimuth of each $(\boldsymbol{y}_i - \boldsymbol{x}_i)$ from the corresponding $\boldsymbol{x}_i$.

As a toy problem, we demonstrate our approach on  $G=\mathrm{SO}(2)$, i.e., the group of rotations of the plane.
In this case, the Lie algebra $\mathfrak{g}$ associated to $G$ has an infinitesimal generator of the form:
\begin{equation}
    \mathbf{A}^{*} =
    \left(
    \begin{array}{cc}
        0 & 1\\
        -1 & 0
    \end{array}
    \right)
    \textrm{, hence }
    \boldsymbol{\phi}:t\mapsto
    \left(
    \begin{array}{cc}
        \cos t & \sin t\\
        -\sin t & \cos t
    \end{array}
    \right)
\end{equation}

We train the neural network on a synthetic dataset of 1k pairs $(\boldsymbol{x}_i, \boldsymbol{y}_i)_i$, where the source points are drawn uniformly at random on the unit circle.
Each angle $t_i$ between the source point $\boldsymbol{x}_i$ and the destination point $\boldsymbol{y}_i$ is drawn uniformly at random in $]0, \frac{\pi}{30}]$.

\begin{figure}
    \centering
    \includegraphics[width=7.25cm]{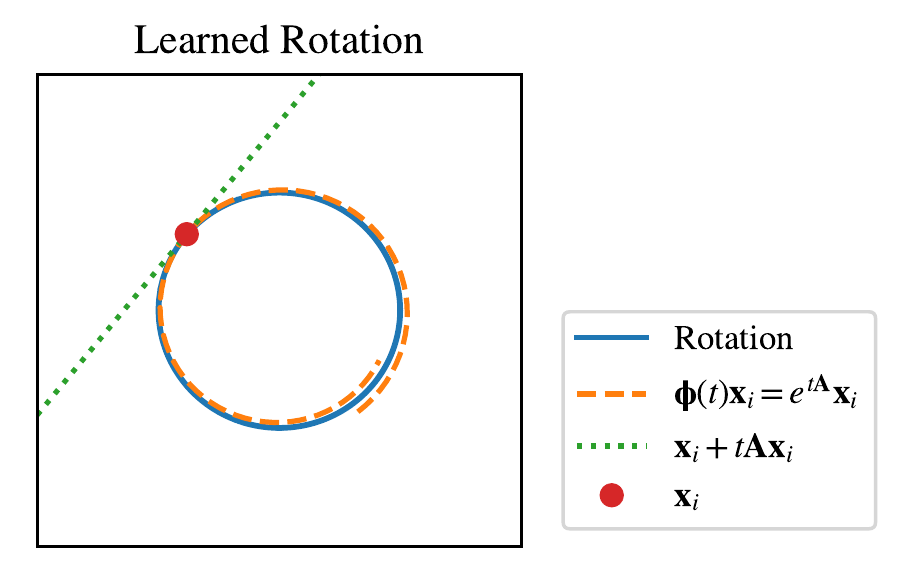}
    \caption{
    Reconstructing rotation using a learned Lie algebra from data pairs.}
    \label{fig:circle}
\end{figure}

Figure \ref{fig:circle} illustrates our result with a specific source point $\boldsymbol{x}_i$ (red dot).
As expected, we observe that the vector $\mathbf{A}\boldsymbol{x}_i$ (dashed green line) is tangent to the orbit of $G$ (blue circle).
Furthermore, the locus of $\boldsymbol{\phi}(t)\boldsymbol{x}_i$ for $t\in[0,2\pi]$ (dashed orange curve) is almost circular.

\section{Conclusion}
We have presented a simple method for learning the Lie algebra associated to any one-parameter subgroup of $\mathrm{GL}_n (\mathbb{R})$.
This method has the advantage of not requiring any human intervention other than collecting data samples by pairs, i.e., before and after the action of the group.
Future work will strive to extend the scope of our method to real-world factors of variability, such as pitch shifts in music or viewpoints in computer vision.
As a long-term goal, it would be interesting to employ the proposed method as a pre-training stage in representation learning, thus yielding convolutional operators whose invariance properties are automatically adapted to the dataset at hand.
\end{document}